\documentclass{article}

\PassOptionsToPackage{numbers}{natbib}

\usepackage[final]{neurips_2025}

\usepackage{changepage}
\usepackage[utf8]{inputenc} 
\usepackage[T1]{fontenc}    
\usepackage{hyperref}       
\usepackage{url}            
\usepackage{booktabs}       
\usepackage{amsfonts}       
\usepackage{nicefrac}       
\usepackage{microtype}      
\usepackage{xcolor}         
\usepackage{graphicx}
\usepackage{caption}
\usepackage{xspace}
\usepackage{enumitem}
\usepackage{array}
\usepackage[table]{xcolor}
\usepackage{multirow} 
\usepackage{tabularx}
\usepackage{lineno}
\usepackage{amsmath}
\usepackage{adjustbox}
\usepackage{hyperref}
\usepackage{cleveref}

\title{Multimodal OCR: Parse Anything from Documents}

\author{
\textbf{Handong Zheng}$^1\thanks{Core contribution}$ \quad 
\textbf{Yumeng Li}$^{2}$\footnotemark[1] \quad \textbf{Kaile Zhang}$^1$\footnotemark[1] \quad \textbf{Liang Xin}$^2$\footnotemark[1] \quad \textbf{Guangwei Zhao}$^2$  \\ \quad \textbf{Hao Liu}$^2$  
\quad \textbf{Jiayu Chen}$^{2}$ \quad \textbf{Jie Lou}$^{2}$ \quad \textbf{Qi Fu}$^{2}$  \quad \textbf{Rui Yang}$^{2}$ \quad \textbf{Shuo Jiang}$^{2}$  \\
\quad \textbf{Weijian Luo}$^{2}$  \quad \textbf{Weijie Su}$^{2}$ \quad \textbf{Weijun Zhang}$^{2}$  \quad
\textbf{Xingyu Zhu}$^{2}$   \quad \textbf{Yabin Li}$^{2}$ \\ \quad \textbf{Yiwei Ma}$^{2}$ \quad \textbf{Yu Chen}$^{2}$ \quad \textbf{Yuqiu Ji}$^{2}$
\quad \textbf{Zhaohui Yu}$^{2}$ 
\quad \textbf{Guang Yang}$^{2}\thanks{Project leader}$ \\ \quad
\textbf{Colin Zhang}$^{2}\thanks{Corresponding author}$  \quad \textbf{Lei Zhang}$^{2}$  \quad \textbf{Yuliang Liu}$^{1\dagger\ddagger}$ \quad \textbf{Xiang Bai}$^{1}$\footnotemark[3]\\
\\
$^1$Huazhong University of Science and Technology \quad $^2$hi lab, Xiaohongshu Inc 
}

\usepackage{multirow}
\usepackage{booktabs}
\usepackage{multirow}
\usepackage{pifont} 
\usepackage{listings}

\newcommand{\cmark}{\ding{51}} 
\newcommand{\xmark}{\ding{55}} 
\newcommand{\pmark}{\raisebox{0.15em}{\scriptsize$\triangle$}}
\begin{document}

\maketitle

\begin{abstract}
We present Multimodal OCR (MOCR), a document parsing paradigm that jointly parses text and graphics into unified textual representations. Unlike conventional OCR systems that focus on text recognition and leave graphical regions as cropped pixels, our method, termed dots.mocr, treats visual elements such as charts, diagrams, tables, and icons as first-class parsing targets, enabling systems to parse documents while preserving semantic relationships across elements. It offers several advantages: (1) it reconstructs both text and graphics as structured outputs, enabling more faithful document reconstruction; (2) it supports end-to-end training over heterogeneous document elements, allowing models to exploit semantic relations between textual and visual components; and (3) it converts previously discarded graphics into reusable code-level supervision, unlocking multimodal supervision embedded in existing documents. To make this paradigm practical at scale, we build a comprehensive data engine from PDFs, rendered webpages, and native SVG assets, and train a compact 3B-parameter model through staged pretraining and supervised fine-tuning. We evaluate dots.mocr from two perspectives: document parsing and structured graphics parsing. On document parsing benchmarks, it ranks second only to Gemini 3 Pro on our OCR Arena Elo leaderboard, surpasses existing open-source document parsing systems, and sets a new state of the art of 83.9 on olmOCR Bench. On structured graphics parsing, our model achieves higher reconstruction quality than Gemini 3 Pro across image-to-SVG benchmarks, demonstrating strong performance on charts, UI layouts, scientific figures, and chemical diagrams. These results show a scalable path toward building large-scale image-to-code corpora for multimodal pretraining. Code and models are publicly available at \url{https://github.com/rednote-hilab/dots.mocr}.
\end{abstract}

\begin{figure}[hpt]
    \centering
    \includegraphics[width=0.94\linewidth]{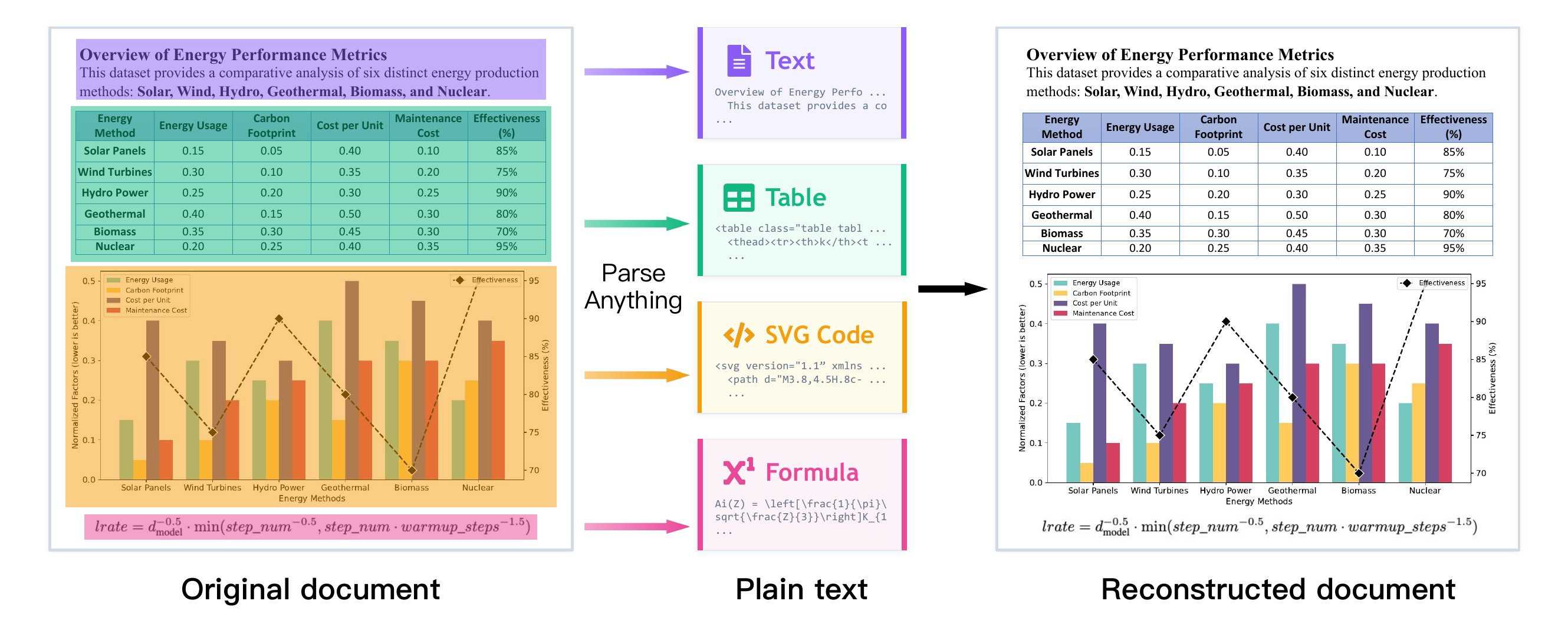}
    \caption{Overview of MOCR. Given a document image, MOCR parses anything on the page into unified, ordered textual representations, capturing both textual and visual structures and enabling faithful reconstruction of the original document.}
    \label{fig:main}
\end{figure}

\section{Introduction}
\label{sec:intro}
In the era of large language and multimodal models, document parsing has become a core data engine for pretraining and retrieval because it determines how much reliable, structured supervision can be recovered from the massive volume of PDFs, scans, and screenshots that store real-world knowledge~\cite{huang2022layoutlmv3}. However, documents convey information not only through text but also through graphics such as charts, diagrams, flow charts, UI elements, and scientific illustrations. Existing document parsing pipelines remain largely text-centric: they focus on recognizing and organizing textual content while treating non-textual elements as figure regions that are simply cropped as pixels~\cite{ouyang2025omnidocbench}. As a result, much of the structural and semantic information encoded in document graphics is discarded, making current document parsing inherently lossy and limiting the amount of supervision that can be extracted from documents~\cite{li2025monkeyocr,niu2025mineru2}.

Recent advances in vision-language models make it increasingly feasible to recover structured representations from document visuals rather than preserving them as pixels. Beyond captioning, modern models show a growing capability to generate executable representations conditioned on images, enabling the reconstruction of underlying structures from visual inputs. Early work on translating UI screenshots into code (e.g., pix2code) explored this direction, and more recent approaches extend it to richer program spaces such as SVG, where images can be converted into renderable vector code~\cite{beltramelli2018pix2code,rodriguez2025starvector}. These developments suggest that document parsing can move beyond text extraction and instead aim to recover all information-bearing elements in documents as structured outputs.

Motivated by this observation, we introduce \textbf{Multimodal OCR (MOCR)}, a document parsing paradigm that aims to \textit{parse anything} in documents, including text, layout structures, tables, and information-dense graphics such as charts, diagrams, icons, and UI components, as illustrated in Fig.~\ref{fig:main}. Unlike traditional OCR pipelines that primarily recover text while leaving graphical regions as raster crops (Fig.~\ref{fig:compare}), MOCR treats both textual and visual elements as first-class parsing targets and converts them into reusable structured outputs. In particular, document graphics are represented as renderable SVG code together with textual content, allowing charts, diagrams, and other visual elements to be reconstructed as structured representations that can serve as reusable supervision for downstream reasoning and multimodal pretraining.

While MOCR provides a unified paradigm for parsing both textual and graphical elements, making it scalable remains challenging. First, supervision for graphics is scarce, as real documents rarely provide aligned program representations for visual elements. Second, renderable programs are inherently non-unique since different codes can produce visually identical outputs, requiring normalization and quality control during training. Third, the task demands precise visual grounding together with long-sequence structured generation, which is substantially more difficult than text-only OCR.

To address these challenges, we develop a scalable system, termed dots.mocr, trained with a large data engine spanning PDFs, rendered webpages, and native SVG graphics. Our training follows a staged recipe that combines large-scale OCR supervision with graphics-centric signals from naturally structured sources while applying normalization and quality control to align predicted code with faithful rendering. This design enables our method to generalize across both traditional document parsing and structured graphics reconstruction within a single unified architecture.

\begin{figure}[t!]
    \centering
    \includegraphics[width=0.90\linewidth]{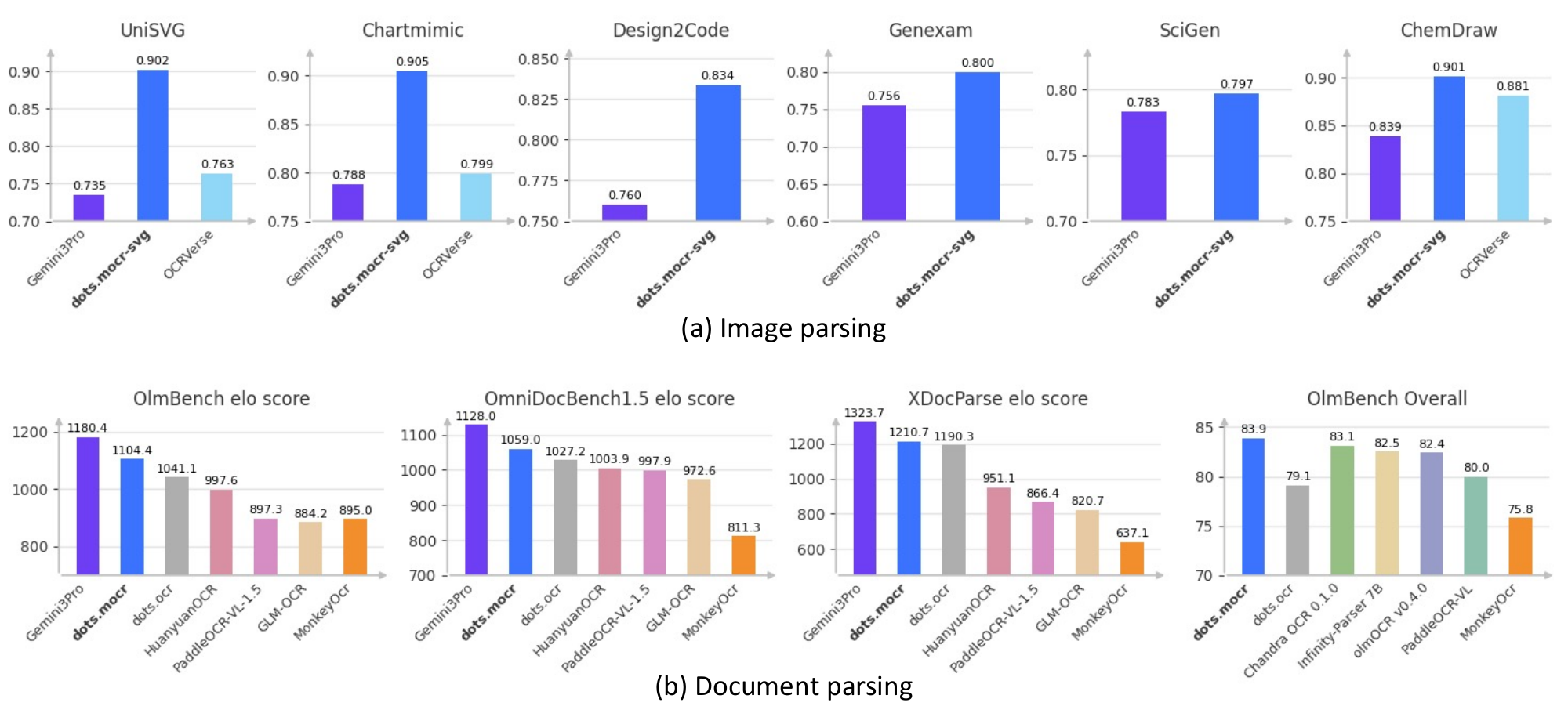}
    \caption{Overall performance comparison results. 
(a) reports metrics for image parsing tasks within documents. 
dots.mocr-svg are further enhanced with additional task-specific training for graphics parsing. (b) reports metrics for general document parsing tasks (olmOCR-Bench, OmniDocBench 1.5, and XDocParse). 
}
    \label{fig:metric}
\end{figure}

The main advantages of this work are summarized as follows:

\begin{itemize}
\item We introduce MOCR, a generalized OCR paradigm that elevates visual symbols to first-class parsing targets and recovers document graphics as reusable, renderable code rather than raster crops, unlocking a new source of structured supervision from existing documents.
\item Our system, dots.mocr, proposes a unified learning formulation that makes it practical at scale under sparse and non-unique program supervision through normalization and training stabilization that align generated code with rendering fidelity, supported by a vision encoder trained entirely from scratch.
\item dots.mocr demonstrates strong and balanced performance across document parsing and structured graphics reconstruction (Fig.~\ref{fig:metric}), ranking second only to Gemini 3 Pro on the OCR Arena Elo leaderboard, setting a new state-of-the-art on olmOCR-Bench, surpassing Gemini 3 Pro on image-to-SVG benchmarks, and maintaining strong visual grounding and reasoning performance on OCRBench beyond parsing, all within a compact 3B-parameter model.
\end{itemize}

\section{Related Work}
\label{sec:related}

\subsection{Text Parsing}
   Text parsing methods have grown rapidly in recent years, aiming to extract and analyze textual content from diverse document formats, including PDFs, web pages, slides, spreadsheets, scanned documents, and scene-text images~\cite{ke2025large}. Existing approaches largely fall into three categories depending on whether and how they leverage vision–language models (VLMs). Traditional systems typically follow a multi-stage pipeline with layout analysis, detection, recognition, and reading-order prediction, as exemplified by pp-structurev3~\cite{cui2025paddleocr3}, which offers modular designs and deployment-oriented integrations; however, they can accumulate errors across stages. A second line augments such pipelines with VLM components to strengthen semantic reasoning while retaining explicit structure, including MonkeyOCR~\cite{li2025monkeyocr}, MinerU 2.5~\cite{niu2025mineru2}, and PaddleOCR-VL~\cite{cui2025paddleocr}; these hybrids improve understanding in many settings yet remain primarily text-focused and inherit some pipeline complexity. Finally, end-to-end VLM-based models cast parsing as direct visual-to-text generation, such as DeepSeek-OCR~\cite{wei2025deepseek}, GOT-OCR~\cite{wei2024general}, and OCRVerse~\cite{zhong2026ocrverse}, achieving strong cross-domain generalization via large-scale pretraining, while still facing challenges in maintaining faithful structure under dense layouts (e.g., tables and formulas). 

\subsection{Structured Graphics Parsing}
    Structured graphics parsing extends text parsing to the recovery of layout, geometry, and styling cues such as shapes, lines, and spatial relations, aiming to translate images into executable, renderable representations (for example, HTML, LaTeX, SVG, or Python) rather than character-level transcripts. Website and UI parsing exemplify this direction by converting screenshots into DOM-like structures or front-end code: Pix2Struct~\cite{lee2023pix2struct} pretrains vision-to-text translation for simplified HTML from masked webpage images, Design2Code~\cite{si2025design2code} benchmarks screenshot-to-implementation generation and highlights persistent fidelity gaps, and OmniParser~\cite{lu2024omniparser} extracts UI elements directly from pixels. Beyond HTML, target languages are often domain-driven, with Plot2Code~\cite{wu2025plot2code}, ChartMimic~\cite{yang2024chartmimic}, and ChartMaster~\cite{tan2025chartmaster} reconstructing charts via programmatic Python rendering, and ChemDraw-style settings mapping molecular diagrams to structured strings such as SMILES~\cite{zhao2025vincicoder}. SVG has emerged as a particularly expressive target because it explicitly encodes geometry and style; recent methods translate images into SVG for icons and vector graphics, including StarVector~\cite{rodriguez2025starvector}, OmniSVG~\cite{yang2025omnisvg}, and UniSVG~\cite{li2025unisvg}. While broader unification efforts such as OCRVerse~\cite{zhong2026ocrverse} combine OCR, chart parsing, SVG reconstruction, web layout generation, and other structured targets within a single vision–language model through prompting, a persistent challenge is matching specialized systems on individual tasks while maintaining robust generalization to complex structured images.

\begin{figure}[t!]
    \centering
    \includegraphics[width=0.90\linewidth]{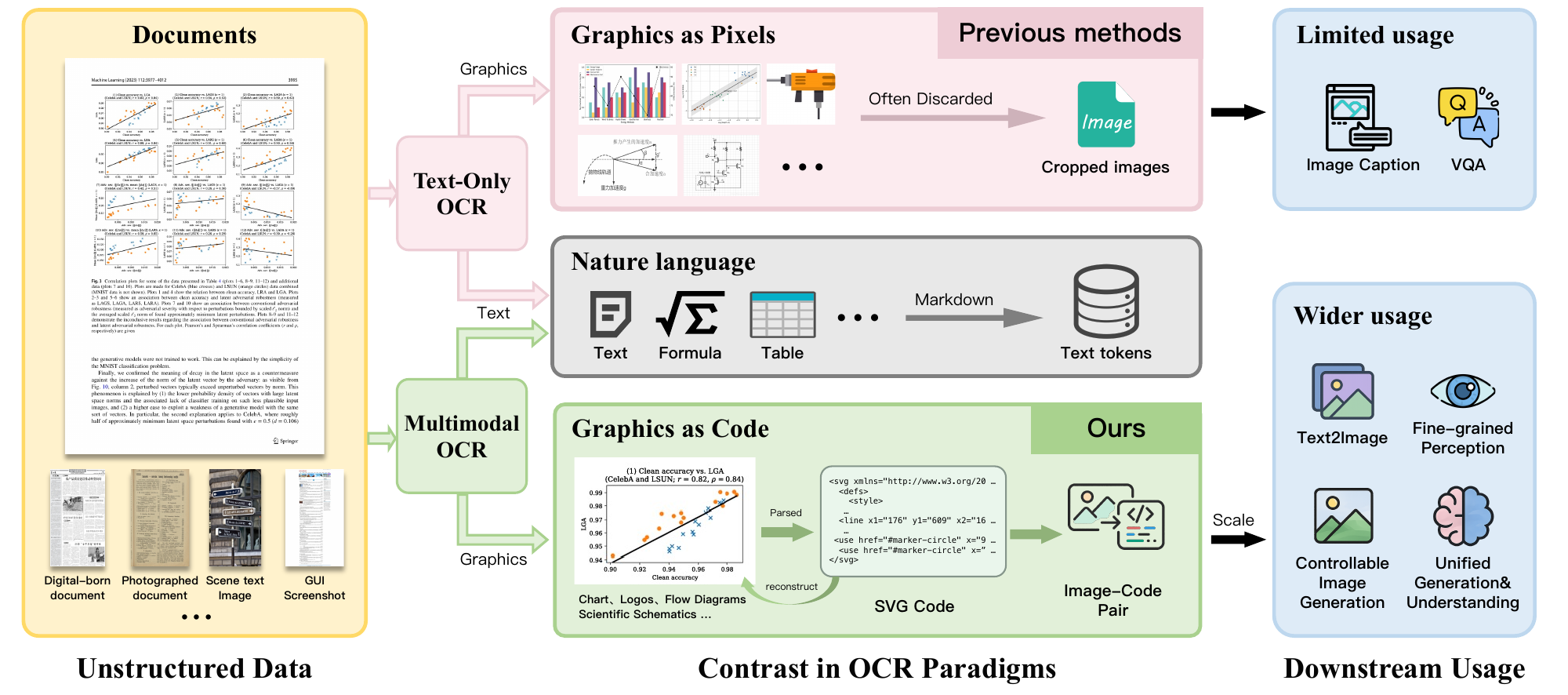}
    \caption{Comparison between traditional text-only OCR and MOCR paradigms. Traditional OCR treats graphics as pixels and often discards them, while MOCR parses graphics into structured code (e.g. SVG), enabling faithful reconstruction and broader downstream applications.}
    \label{fig:compare}
\end{figure}

In this context, we introduce MOCR to parse anything in documents, converting not only text but also charts, diagrams, UI elements, icons, and domain drawings into reusable, renderable representations instead of raster crops. MOCR aims to reframe document parsing as a scalable source of executable supervision for multimodal pretraining and retrieval, bridging text-focused parsers and task-specific graphic systems. An intuitive comparison among different systems is shown in Tab.~\ref{tab:pipeline-comparion} and Fig.~\ref{fig:compare}.

\begin{table}[!t]
    \small
    \centering
\caption{Comparison of different VLMs and OCR systems across multiple tasks.
\cmark\ indicates Supported (good), \pmark\ indicates Supported but Underperforming, and \xmark\ indicates Not Supported.}
    \setlength{\abovecaptionskip}{0.35cm}
    \setlength{\tabcolsep}{0.55mm}{
        \begin{tabular}{l|cc|cccccc}
            \toprule
            \multirow{2}{*}{\parbox{1.6cm}{\centering \textbf{Model\\Type}}} &
            \multirow{2}{*}{\parbox{1.9cm}{\centering \textbf{Inference\\Type}}} &
            \multirow{2}{*}{\parbox{2.2cm}{\centering \textbf{Model\\Name}}} & \multicolumn{6}{c}{\textbf{Task}}\\
            &&&  Layout-Det & Parsing & Spotting & VQA & IE & Graphics \\
            \midrule
            \multirow{3}{*}{\parbox{1.6cm}{\centering \textbf{Cascade Pipeline}}}&
            \multirow{3}{*}{\parbox{1.6cm}{\centering \textbf{Multi-Step}}}&
            PaddleOCR-V5~\cite{cui2025paddleocr3}& \xmark&\xmark& \cmark& \xmark & \xmark & \xmark
            \\
            &&Marker-1.8.2& \cmark&\cmark& \xmark& \xmark & \xmark & \xmark
            \\
            &&PP-StructureV3~\cite{cui2025paddleocr3}& \cmark&\cmark& \xmark & \xmark & \xmark& \xmark
            \\
            \midrule
            \multirow{4}{*}{\parbox{1.6cm}{\centering \textbf{Specialized VLMs (Modular)}}}&
            \multirow{4}{*}{\parbox{1.6cm}{\centering \textbf{Multi-step}}}&
            MonkeyOCR-pro-3B~\cite{li2025monkeyocr}& \cmark&\cmark& \xmark& \xmark & \xmark & \xmark
            \\
            &&MinerU2.5~\cite{niu2025mineru2}& \cmark&\cmark& \xmark& \xmark & \xmark & \xmark
            \\
            &&PaddleOCR-VL-1.5~\cite{cui2026paddleocr}& \cmark&\cmark& \xmark& \xmark & \xmark & \xmark
            \\
 & & GLM-OCR~\cite{duan2026glmocrtechnicalreport}& \cmark& \cmark& \xmark& \xmark & \cmark&\xmark
\\
            \midrule
            \multirow{2}{*}{\parbox{1.6cm}{\centering \textbf{Documents VLMs}}}&
            \multirow{2}{*}{\parbox{1.6cm}{\centering \textbf{One-Step}}}&
            Text-Monkey~\cite{liu2026textmonkey}& \xmark&\xmark & \cmark & \cmark & \cmark & \xmark
            \\
            &&mPLUG-DocOwl2~\cite{hu2025mplug}& \xmark&\cmark & \cmark & \cmark & \cmark & \xmark
            \\
            \midrule
            \multirow{2}{*}{\parbox{1.6cm}{\centering \textbf{General VLMs}}}&\multirow{2}{*}{\parbox{1.6cm}{\centering \textbf{One-Step}}}&Qwen3-VL-235B-Instruct~\cite{bai2025qwen3}& \cmark&\cmark & \cmark & \cmark & \cmark & \pmark
            \\
            &&Gemini 3 pro~\cite{google2025gemini3} & \cmark&\cmark & \cmark & \cmark & \cmark & \pmark
            \\
            \midrule
            \multirow{5}{*}{\parbox{1.6cm}{\centering \textbf{Specialized VLMs (End2End)}}}&
            \multirow{5}{*}{\parbox{1.6cm}{\centering \textbf{One-Step}}}
            &dots.ocr~\cite{li2025dots}& \cmark&\cmark& \xmark& \xmark & \xmark & \xmark
            \\
            &&HunyuanOCR~\cite{team2025hunyuanocr}& \xmark&\cmark & \cmark & \cmark & \cmark & \xmark\\
            &&DeepSeek-OCR2~\cite{wei2026deepseek}& \cmark&\cmark& \xmark& \xmark& \xmark& \xmark\\
            & & OCRVerse~\cite{zhong2026ocrverse}& \xmark & \cmark & \xmark& \xmark& \xmark&\pmark
\\
             & & dots.mocr&             \cmark & \cmark & \cmark & \cmark & \cmark &\cmark
\\
            \bottomrule
        \end{tabular}}
    \label{tab:pipeline-comparion}
\end{table}

\section{Multimodal OCR}
\label{sec:method}
    MOCR is designed by unifying page-level parsing tasks within a single model, including document parsing, webpage and UI parsing, scene-text parsing, and structured graphics parsing. This unification turns documents and screens into a richer data engine by recovering not only text but also visual symbols as reusable, renderable code (e.g., SVG) that is executable, editable, and compositional, enabling scalable supervision for pretraining and retrieval beyond raster crops.

\subsection{Task Definition}
\label{sec:task_definition}
        MOCR aims for the comprehensive parsing of document pages—including PDF renderings, digital scans, webpages, and scene-text images. Unlike traditional text-centric pipelines that treat non-textual elements as inert raster crops, MOCR treats both text and visual symbols as first-class parsing targets. This approach explicitly recovers information-dense graphics—such as charts, diagrams, icons, and schematics—into structured, reusable representations, thereby transforming static pixels into actionable data for downstream reasoning and multimodal training.
    
    Given an input image $\mathbf{I}$, the task is to generate an ordered sequence of parsed elements $\mathbf{S}$:
        \begin{equation}
        \mathbf{S} = \big[(\mathcal{B}_1, c_1, p_1), \ldots, (\mathcal{B}_K, c_K, p_K)\big].
        \end{equation}
    Where each constituent element is defined by:
    $\mathcal{B}_k$, the spatial region or bounding box. $c_k$, the semantic category or element type. $p_k$, the associated payload. The sequence $\mathbf{S}$ is generated following a human-centric reading order, allowing the model to encode structural hierarchies and logical relations implicitly through the generation sequence and specialized delimiters, rather than relying on an external relation module.
    
    The payload $p_k$ is a type-specific serialization of the content within region $\mathcal{B}_k$, determined by the semantic type $c_k$.
    For text-centric regions (e.g., text lines/blocks, tables, and formulas), $p_k$ corresponds to their transcriptions in appropriate symbolic forms, such as plain text, table markup, or \LaTeX{}. For visual symbols that admit a concise, programmatic description—such as UI components, icons, and charts—$p_k$ is a renderable structured representation, i.e., image-to-SVG conversion. By parsing eligible graphics into SVG code, MOCR facilitates ``render-and-reuse'' workflows. Notably, complex real-world imagery or natural photographs, which lack a compact programmatic description, are retained as raster content. This strategic shift enables documents to contribute not only textual tokens but also granular, controllable structural supervision for the next generation of multimodal pretraining.

    In the current release, MOCR is task-conditioned and does not yet produce a single one-pass output that simultaneously includes full-page document parsing and visual-symbol (e.g., SVG) parsing; instead, we obtain a complete multimodal parse by running page-level text parsing and region-level image-to-SVG decoding in separate passes.
     
\subsection{Model Architecture}
\label{sec:model_architecture}
The architecture of our method adheres to the fundamental design principles established in previous work~\cite{li2025dots}. It comprises three primary components: a high-resolution vision encoder, a lightweight multimodal connector, and an autoregressive language model (LLM) decoder.

\paragraph{High-Resolution Vision Encoder.}
    The vision encoder is a 1.2B-parameter backbone trained entirely from scratch, which ensures the encoder develops feature representations natively optimized for document parsing, enabling the joint modeling of dense text and geometry-sensitive visual symbols (e.g., charts, diagrams, and schematics). Architecturally, the encoder is engineered to ingest native high-resolution inputs of up to $\sim$11M pixels. This high-capacity throughput is essential for preserving fine-grained details and maintaining long-range spatial coherence across a full page. Such resolution is critical not only for legibility in small-font text or dense layouts but also for the precise perception of graphic primitives—such as chart markers and diagrammatic strokes—which must be accurately localized to be recovered as structured code.

\paragraph{Structured Language Decoder.}
    For the autoregressive decoder, we use Qwen2.5-1.5B. The key consideration is the capacity and cost trade-off for unified MOCR parsing: models substantially smaller than 1.5B often struggle to simultaneously handle heterogeneous page content (text, layout structure, and visual symbols) and generate long, highly structured outputs such as SVG programs within a single autoregressive decoding process, while significantly larger decoders increase training and inference costs. Initializing from a base model (rather than a chat-specialized model) provides a neutral starting point for large-scale pretraining, where the model must learn non-natural, strongly structured target sequences and long-range dependencies as part of the parsing objective.

\subsection{Training Recipe}
\label{sec:training_recipe}
    Our training strategy is intentionally data-driven. Given the broad coverage of MOCR, our goal is not to introduce task-specific optimization heuristics, but to design an efficient curriculum that lowers learning difficulty, stabilizes multi-task joint training, and enables a single model to absorb heterogeneous supervision produced by our data engine.

    We perform large-scale pretraining in three successive stages, each serving a distinct purpose. The first stage establishes a stable vision-language interface through general-purpose vision training so that the language model can reliably consume visual tokens and ground generation on visual inputs. The second stage conducts broad pretraining on a unified mixture of general vision data and text-only document parsing supervision, building strong text-centric parsing foundations while maintaining general visual robustness. The third stage shifts the mixture toward MOCR-specific targets by decreasing the proportion of general vision data and increasing the emphasis on multimodal document parsing, strengthening OCR-centric parsing together with visual-symbol parsing instantiated as image-to-SVG. Across all stages, we keep a single autoregressive objective, predicting structured parsing sequences conditioned on the input image and task instruction while controlling optimization stability via mixture reweighting and curriculum scheduling. We also progressively increase input resolution across stages to match the growing difficulty of dense page parsing and long structured generation.

    After pretraining, we perform instruction tuning using a curated high-quality supervised set constructed by our data engine. Relative to pretraining, this stage prioritizes supervision reliability and task usability: we filter and refine examples to correct systematic errors, align output conventions, and improve end-to-end parsing fidelity across tasks. For visual-symbol parsing, instruction tuning is especially sensitive to target consistency, so SVG-specific handling (e.g., canonicalization, viewBox normalization, and complexity reduction) is treated as part of the data engine, while the training recipe focuses on integrating these refined signals into a stable multi-task SFT mixture. We release two checkpoints with the same pretraining: \textbf{dots.mocr} and \textbf{dots.mocr-svg}, where the latter increases the SVG share and up-weights harder SVG programs during SFT to better prioritize image-to-SVG parsing under the same parameter budget.

\subsection{Data Engine}
\label{sec:data_engine}
    Training a single model for MOCR places unusually strict requirements on the training corpus. Beyond robustness to scripts, diverse layouts, and long-range reading structures, the model must also learn to parse visual symbols (such as charts, diagrams, icons, and schematics) into reusable structured representations rather than leaving them as raster crops. No existing dataset provides this coverage at sufficient scale and quality.

    Our training corpus is built from four complementary sources: (i) PDF documents for text-language page parsing, (ii) web-derived pages rendered into images with aligned structural signals, (iii) native SVG assets for image-to-SVG supervision, and (iv) general-purpose data to maintain broad robustness and downstream usability. We apply lightweight quality control for pretraining to remove obvious noise while preserving diversity, and curate a smaller, higher-precision subset for instruction tuning with stricter verification and convention alignment.

    \textbf{PDF documents.} We construct multilingual document parsing supervision from raw PDFs using dots.ocr as an auto-labeling engine, producing structured page transcriptions with layout regions and reading order. We curate the PDF pool via stratified sampling over language, domain, and layout complexity (estimated by lightweight proxies such as block count, text density, and the presence of tables/formulas) to emphasize hard regimes. For instruction tuning, we further improve reliability through (i) verification with rule-based sanity checks and render-based comparison against the input page, and (ii) distillation that relabels or filters samples with stronger supervision to correct common errors.

    \textbf{Webpages.} We crawl and render webpages into page images and convert them into the same MOCR parsing format as PDFs. This source broadens the distribution with naturally high-resolution and complex layouts, provides aligned structural signals from HTML/DOM to reduce label noise, and supplies abundant SVG-native icons, charts, and diagrams that further support visual-symbol parsing.

    \textbf{SVG graphics.} A central goal of MOCR is to parse eligible graphics into reusable, renderable representations rather than keeping them as raster crops. Since many icons, charts, and UI graphics on the web are natively stored as SVG, we collect such assets from diverse sources and render them to construct image–SVG pairs. Our pipeline consists of two stages: cleaning and sampling. During cleaning, we use \texttt{svgo}~\footnote{https://github.com/svg/svgo} to remove irrelevant metadata, normalize numeric precision, and standardize code structure, followed by deduplication at both the code and image levels using textual matching and perceptual hashing (pHash) on rendered images. During sampling, we perform domain-level balancing to avoid over-representation from individual sources and apply complexity-aware sampling based on SVG program complexity to maintain a balanced mix of simple and complex graphics.

    \textbf{General-purpose data.} We additionally include generic vision and OCR supervision (e.g., grounding and counting) to preserve broad capabilities alongside page-level parsing.

    This data engine enables unified training over text parsing and visual-symbol parsing, converting previously raster-only graphics into reusable program supervision for MOCR.

\subsection{Automated Evaluation via OCR Arena}
\label{sec:elo}
Traditional metrics such as Word Error Rate (WER) or Normalized Edit Distance (NED), as well as structure-aware scores like TEDS~\cite{zhong2020image} for tables and CDM~\cite{wang2025image} for formulas, often fail to reflect the true end-to-end quality of complex Markdown OCR outputs because they rely on rule-based matching to ground truth and are sensitive to non-unique but semantically equivalent serializations~\cite{horn2025benchmarking}. To address this limitation, we adopt an automated evaluation framework based on the LLM-as-a-Judge paradigm, referred to as \textit{OCR Arena}. In this framework, a high-capacity vision-language model, such as Gemini 3 Flash, evaluates pairs of model outputs given the original document image and their generated Markdown transcriptions, judging which result better preserves fidelity, structure, and formatting or declaring a tie when both are comparable.

To ensure the integrity of our benchmarking and mitigate the well-documented issue of positional bias—where LLMs tend to favor the candidate presented first—we employ a rigorous symmetric evaluation protocol. Every pairwise comparison between Model $A$ and Model $B$ is conducted in two distinct trials: one where Model $A$ is presented as the first candidate, and another where the presentation order is reversed. A model is credited with a victory only if the judge’s decision remains consistent across both trials. If the judge's preference shifts solely based on the presentation order, or if the results are contradictory, the battle is categorized as inconsistent and treated as a tie. This dual-trial approach effectively filters out presentation artifacts and ensures that the final rankings reflect genuine model superiority.

To synthesize the outcomes of thousands of pairwise battles into a unified and interpretable leaderboard, we utilize the Elo rating system, a probabilistic ranking method originally developed for competitive games. Each model begins with an initial rating $R$, and for any given battle between Model $A$ and Model $B$, the expected probability of Model $A$ winning is calculated using the logistic curve:
\begin{equation}
E_A = \frac{1}{1 + 10^{(R_B - R_A) / 400}}
\end{equation}
Following the conclusion of a battle, the ratings are updated according to the actual outcome $S_A$ (where $S_A=1$ for a win, $0.5$ for a tie, and $0$ for a loss) using the update rule:
\begin{equation}
R'_A = R_A + K \cdot (S_A - E_A)
\end{equation}
The factor $K$ represents the sensitivity of the rating update, which we set to 32. This framework allows for a dynamic and scalable ranking that accounts for the relative strength of opponents, ensuring that victories against stronger models are weighted more heavily than those against weaker ones.

Given that the final Elo ratings can be influenced by the specific sequence in which battles are processed, we incorporate bootstrap resampling to enhance statistical robustness. We perform 1,000 iterations of the Elo calculation, with the entire battle history randomly shuffled in each round. The final reported Elo score for each model is the mean value derived from this distribution. This methodology guarantees that our leaderboard is stable and accurately represents the comparative performance of the OCR models across the entire evaluation dataset, providing a reliable measure of progress in high-fidelity document transcription.

\section{Experiments}
\label{sec:experiments}

\subsection{Document Parsing}
Existing document parsing benchmarks often rely on strict matching rules over the predicted structured text, which can be sensitive to surface-form differences even when the underlying parsing is acceptable. To complement such evaluations, we adopt an Elo-style paired comparison protocol (introduced in the previous section), where model outputs are judged pairwise and aggregated into Elo ratings.
In our setup, the Elo evaluation is conducted using Gemini 3 Flash as the judge, and all models in Tab.~\ref{tab:model_compare} are evaluated under the same protocol, making the Elo scores directly comparable.

\begin{table}[h]
\centering
\caption{Elo comparison on olmOCR-Bench~\cite{poznanski2025olmocr}, OmniDocBench(v1.5)~\cite{ouyang2025omnidocbench}, and XDocParse~\cite{li2025dots}. All models are evaluated under the same Elo protocol with Gemini 3 Flash as judge, on benchmark-provided document images.}
\resizebox{0.60\linewidth}{!}{
\begin{tabular}{lcccl}
\toprule
\textbf{Models} & \textbf{olmOCR-Bench} & \textbf{OmniDocBench1.5} & \textbf{XDocParse}  &\textbf{Average}\\
\midrule
Gemini 3 Pro~\cite{google2025gemini3}         & \textbf{1180.4}& \textbf{1128.0}& \textbf{1323.7}& \textbf{1210.7}\\
\rowcolor{gray!20}
dots.mocr       & 1104.4& 1059.0& 1210.7&1124.7\\
dots.ocr~\cite{li2025dots} & 1041.1& 1027.2& 1190.3&1086.2\\
HunyuanOCR~\cite{team2025hunyuanocr}        &  997.6&  1003.9&  951.1&984.2\\
PaddleOCR-VL-1.5~\cite{cui2026paddleocr}   &  897.3&  997.9&  866.4&920.5\\
GLM-OCR~\cite{duan2026glmocrtechnicalreport}            &  884.2&  972.6&  820.7&892.5\\
MonkeyOCR-pro-3B~\cite{li2025monkeyocr}            &  895.0&  811.3&  637.1&781.1\\

\bottomrule
\end{tabular}}
\label{tab:model_compare}
\end{table}

Across all three benchmarks, dots.mocr achieves the strongest Elo performance among the open-source models listed in the table, indicating consistently strong text-language parsing quality under the same Elo protocol.
Gemini 3 Pro ranks first in all three benchmarks in this comparison.

\begin{table}[h]
\centering
\caption{Performance comparison on olmOCR-Bench.}
\resizebox{\linewidth}{!}{
\begin{tabular}{lccccccccc}
\toprule
\textbf{Model} &
\textbf{ArXiv} &
\textbf{Old scans math} &
\textbf{Tables} &
\textbf{Old scans} &
\textbf{Headers \& footers} &
\textbf{Multi column} &
\textbf{Long tiny text} &
\textbf{Base} &
\textbf{Overall} \\
\midrule
MinerU 2.5.4~\cite{niu2025mineru2}          & 76.6 & 54.6 & 84.9 & 33.7 & \textbf{96.6} & 78.2 & 83.5 & 93.7 & 75.2$\pm$1.1 \\
MonkeyOCR-pro-3B~\cite{li2025monkeyocr}   & 83.8 & 68.8 & 74.6 & 36.1 & 91.2 & 76.6 & 80.1 & 95.3 & 75.8$\pm$1.0 \\
dots.ocr~\cite{li2025dots}               & 82.1 & 64.2 & 88.3 & 40.9 & 94.1 & 82.4 & 81.2 & 99.5 & 79.1$\pm$1.0 \\
DeepSeek-OCR~\cite{wei2025deepseek}           & 77.2 & 73.6 & 80.2 & 33.3 & 96.1 & 66.4 & 79.4 & 99.8 & 75.7$\pm$1.0 \\
Nanonets-OCR2-3B~\cite{Nanonets-OCR2}       & 75.4 & 46.1 & 86.8 & 40.9 & 32.1 & 81.9 & \textbf{93.0} & 99.6 & 69.5$\pm$1.1 \\
PaddleOCR-VL~\cite{cui2025paddleocr}           & 85.7 & 71.0 & 84.1 & 37.8 & 97.0 & 79.9 & 85.7 & 98.5 & 80.0$\pm$1.0 \\
Infinity-Parser 7B~\cite{wang2025infinity}    & 84.4 & 83.8 & 85.0 & 47.9 & 88.7 & 84.2 & 86.4 & 99.8 & 82.5$\pm$? \\
olmOCR v0.4.0~\cite{poznanski2025olmocr}          & 83.0 & 82.3 & 84.9 & 47.7 & 96.1 & 83.7 & 81.9 & 99.7 & 82.4$\pm$1.1 \\
\rowcolor{gray!20}
dots.mocr           & \textbf{85.9} & \textbf{85.5} & \textbf{90.7} & \textbf{48.2} & 94.0 & \textbf{85.3} & 81.6 & 99.7 & \textbf{83.9}$\pm$\textbf{0.9} \\
\bottomrule
\end{tabular}}
\label{tab:doc_subset_compare}
\end{table}

We note that Elo ratings are not expected to numerically align with the strict-match metrics or leaderboard scores reported by existing benchmarks.
First, our Elo protocol relies on pairwise judging rather than surface-form string matching, which mitigates known brittleness in rule-based evaluation (e.g., over-penalizing minor formatting or normalization differences) and can therefore shift absolute rankings.
Second, the benchmark-provided image sets contain relatively fewer formula- and table-heavy pages; since Elo aggregates per-sample pairwise comparisons, these pages receive a proportionally smaller effective weight, consistent with our qualitative observations.
In practice, models with lower Elo ratings often lose even on visually simple pages due to small but salient character-level mistakes (e.g., a single wrong symbol), which are consistently judged as worse in head-to-head comparisons.
Overall, Elo provides complementary signals beyond strict matching by capturing error severity and robustness on common document pages.

To further analyze performance across document regimes, Tab.~\ref{tab:doc_subset_compare} reports a category breakdown on olmOCR-Bench.
dots.mocr achieves the best overall score among all reported systems, and it also attains the highest scores on ArXiv, Old scans math, Tables, and Multi column.
For other categories (e.g., Old scans, Headers \& footers, Long tiny text, and Base), the best scores are achieved by other systems in the table, suggesting remaining headroom in these regimes.

\subsection{Structured Graphics Parsing}
We evaluate structured graphics parsing on a diverse set of benchmarks covering key visual domains: general vector graphics (UniSVG~\cite{li2025unisvg}); scientific charts (ChartMimic~\cite{yang2024chartmimic}); webpage and UI layouts (Design2Code~\cite{si2025design2code}); exam-style diagrams (GenExam~\cite{wang2025genexam}); scientific figures (SciGen~\cite{lin2026scientific}); and chemistry structure diagrams (ChemDraw~\cite{zhao2025vincicoder}). Together, these datasets span icons, charts, layouts, and domain-specific scientific drawings, enabling a broad evaluation of reconstruction quality.

For each benchmark, we use the original image as input, render the predicted structured code, and compute the ISVGEN score (from UniSVG~\cite{li2025unisvg}) between the rendered result and the original image. This render-and-compare protocol provides a unified reconstruction-based metric across all datasets in Tab.~\ref{tab:unisvg_and_downstream}. We compare with an open-source baseline, OCRVerse, and a strong closed-source model, Gemini 3 Pro. OCRVerse generates different program formats depending on the task (e.g., SVG or Python), while Gemini 3 Pro and our models use SVG outputs; all methods are evaluated consistently through rendering and ISVGEN.

\begin{table}[h]
\centering
\caption{ISVGEN reconstruction scores on UniSVG (low-level, high-level, overall) and downstream visual-language benchmarks (higher is better).}
\resizebox{\linewidth}{!}{
\begin{tabular}{lcccccccc}
\toprule
\textbf{Methods} & 
\textbf{UniSVG}\textsuperscript{\textbf{Low-Level}} &
\textbf{UniSVG}\textsuperscript{\textbf{High-Level}} &
\textbf{UniSVG}\textsuperscript{\textbf{Score}} &
\textbf{ChartMimic} &
\textbf{Design2Code} &
\textbf{GenExam} &
\textbf{SciGen} &
\textbf{ChemDraw} \\
\midrule
OCRVerse~\cite{zhong2026ocrverse}           & 0.632 & 0.852 & 0.763 & 0.799 & -     & -     & -     & 0.881 \\
Gemini 3 Pro~\cite{google2025gemini3}         & 0.563 & 0.850 & 0.735 & 0.788 & 0.760 & 0.756 & 0.783 & 0.839 \\
dots.mocr       & 0.850 & 0.923 & 0.894 & 0.772 & 0.801 & 0.664 & 0.660 & 0.790 \\
\rowcolor{gray!20}
dots.mocr-svg   & \textbf{0.860} & \textbf{0.931} & \textbf{0.902} & \textbf{0.905} & \textbf{0.834} & \textbf{0.800} & \textbf{0.797} & \textbf{0.901} \\
\bottomrule
\end{tabular}}
\label{tab:unisvg_and_downstream}
\end{table}

As shown in Tab.~\ref{tab:unisvg_and_downstream}, dots.mocr-svg achieves the best overall performance across datasets, surpassing OCRVerse by +0.139 on UniSVG overall (0.902 vs.\ 0.763) and outperforming Gemini 3 Pro on all reported downstream benchmarks. The improvements are especially clear on structure-sensitive tasks such as ChartMimic and ChemDraw, while remaining strong on layouts and scientific figures. Given the 3B parameter scale, the base dots.mocr does not fully focus on visual-language parsing; dots.mocr-svg strengthens this ability by training with more visual-language data, and we will continue improving structured graphics understanding in future updates.

\subsection{General VQA Evaluation}

\begin{table}[h]
\centering
\caption{Performance comparison on CharXiv, OCR reasoning, and a suite of document and vision-language benchmarks.}
\resizebox{\linewidth}{!}{
\begin{tabular}{lcccccccccc}
\toprule
\textbf{Model} &
\textbf{CharXiv}\textsuperscript{\textbf{Descriptive}} &
\textbf{CharXiv}\textsuperscript{\textbf{Reasoning}} &
\textbf{OCR Reasoning} &
\textbf{InfoVQA} &
\textbf{DocVQA} &
\textbf{ChartQA} &
\textbf{OCRBench} &
\textbf{AI2D} &
\textbf{CountBenchQA} &
\textbf{RefCOCO} \\
\midrule
Qwen3-VL-2B-Instruct~\cite{bai2025qwen3} & 62.3 & 26.8 & -    & 72.4  & 93.3  & -    & 85.8 & 76.9 & 88.4  & - \\
 Qwen3-VL-4B-Instruct~\cite{bai2025qwen3}& 76.2& 39.7& -& 80.3& 95.3& -& 88.1& 84.1& 84.9&-\\
\rowcolor{gray!20}
dots.mocr        & 77.4 & 55.3 & 22.85 & 73.76 & 91.85 & 83.2 & 86.0 & 82.16 & 94.46 & 80.03 \\
\bottomrule
\end{tabular}}
\label{tab:multi_benchmark_compare}
\end{table}

\textbf{General Capability.} Although dots.mocr is primarily designed for multimodal structured parsing, we further evaluate its general vision-language capability on a suite of benchmarks covering document understanding and broader multimodal reasoning (Tab.~\ref{tab:multi_benchmark_compare}). Overall, dots.mocr remains highly competitive against a strong general-purpose baseline. In particular, it shows clear advantages on CharXiv~\cite{wang2024charxiv} in both descriptive and reasoning settings, suggesting stronger fine-grained text-centric understanding and multimodal reasoning. On downstream document VQA and chart understanding tasks (e.g.,OCR Reasoning~\cite{huang2025ocr}, InfoVQA~\cite{mathew2022infographicvqa}, DocVQA~\cite{mathew2021docvqa}, and ChartQA~\cite{masry2022chartqa}), dots.mocr  achieves consistently strong results, indicating that unifying document parsing with broader VLM training does not compromise general-purpose performance. We also observe solid performance on OCRBench~\cite{liu2024ocrbench}, AI2D~\cite{kembhavi2016diagram}, CountBenchQA~\cite{paiss2023countclip}, and RefCOCO~\cite{yu2016modeling}, demonstrating that dots.mocr maintains broad visual grounding and reasoning abilities beyond parsing.

\section{Qualitative examples}
\label{sec:examples}

We present qualitative results to illustrate the structural understanding and reconstruction capabilities of dots.mocr and dots.mocr-svg across diverse visual domains. Instead of producing raster crops, the models generate structured outputs, including layout partitions and executable SVG code, enabling reusable and machine-readable representations. As shown in Fig.~\ref{fig:example-doc} and~\ref{fig:example-parse}, dots.mocr performs robust layout analysis on heterogeneous documents, handling multilingual pages, complex multi-column layouts, dense tables, mathematical formulas, scanned materials, and handwritten notes. It also parses full-length web screenshots with preserved reading order and structured components, and recognizes scene text in real-world images, demonstrating strong generalization beyond traditional OCR settings. 

Fig.~\ref{fig:example-icon}--\ref{fig:example-subject} further show that dots.mocr-svg converts images into structured SVG code for vector-level reconstruction. The model supports basic icons, diverse statistical charts (e.g., bar, line, scatter, and composite charts), as well as complex cross-disciplinary illustrations in science and design, capturing geometric structure and semantic grouping in an editable and scalable representation. 

As illustrated in Fig.~\ref{fig:example-general}, while retaining strong document parsing and SVG reconstruction capabilities, the model also maintains competitive general-purpose vision-language performance, producing coherent and context-aware responses across documents, charts, UI screenshots, and complex illustrations.

\section{Discussion}
\label{sec:discuss}
Beyond the immediate benchmarks, we believe MOCR introduces a new direction for the broader community. By converting document graphics into image–code pairs, MOCR provides a scalable pipeline for constructing large multimodal pretraining corpora. Each chart parsed into SVG and each diagram converted into structured markup can form faithful image, code, and text triples, creating controllable and perturbable training data at a scale limited only by the number of available documents. Although this work instantiates graphics parsing as image-to-SVG, the MOCR paradigm is representation agnostic. Future extensions could target other program spaces such as TikZ for scientific figures, D3.js for interactive visualizations, CAD formats for engineering drawings, or domain-specific markup for chemical structures and circuit diagrams. Moreover, the ability to parse full webpages with diverse layouts, embedded graphics, and multilingual content substantially expands the pool of available training data beyond traditional PDF-centric corpora.

At the system level, MOCR also suggests new opportunities in data construction and evaluation. Our data engine shows that principled normalization, together with render based verification, can address the challenge of non-unique program targets. Scaling these quality control mechanisms through tighter render verification loops, reward model filtering, and self improving data curation offers a clear path to further improvements. Finally, our OCR Arena framework provides a practical alternative to brittle rule-based metrics, and similar judge-based evaluation protocols may become increasingly important as parsing outputs grow in structural complexity and representational diversity.

\section{Conclusion}
\label{sec:conclusion}
This work introduces Multimodal OCR (MOCR), a new document parsing paradigm that broadens document parsing from text extraction to a structured understanding of all information-bearing elements. Conventional systems typically stop at text and treat graphics, charts, diagrams, and icons as opaque pixel regions that are detected but not structurally interpreted. In contrast, the proposed paradigm treats every information-bearing element as a first-class parsing target and recovers visual symbols as reusable, renderable structured code. The key insight is that the richest supervision in many documents is often visual rather than textual; yet, it has historically been discarded by OCR pipelines. By recovering graphics as executable and re-renderable SVG programs, this work transforms static document pixels into structured supervision that can be reused for reasoning and learning. This capability not only broadens the scope of document parsing, but also increases the amount of usable supervision that can be extracted from large document corpora. More broadly, we hope this work helps shift document parsing from text-centric OCR toward document-native multimodal parsing that captures the full visual language of human knowledge.

{
    \small
    \bibliographystyle{ieeenat_fullname}
    \bibliography{main}
}

\newpage

\clearpage

\setcounter{page}{1}


\appendix
\section{Appendix}

\subsection{Other Benchmarks}
\begin{table}[h]
\centering
\caption{Comparison of general-purpose and specialized VLMs on OmniDocBench(v1.5) and pdf-parse-bench. Lower is better for edit metrics.}

\resizebox{\linewidth}{!}{
\begin{tabular}{llcccc}
\toprule
\textbf{Model Type} & \textbf{Methods} & \textbf{Size} &
\textbf{OmniDocBench(v1.5)}\textsuperscript{\textbf{TextEdit}}$\downarrow$ &
\textbf{OmniDocBench(v1.5)}\textsuperscript{\textbf{Read OrderEdit}}$\downarrow$ &
\textbf{pdf-parse-bench} \\
\midrule
\multirow{3}{*}{GeneralVLMs}
& Gemini 2.5 Pro               & -     & 0.075 & 0.097 & 9.06 \\
& Qwen3-VL-235B-A22B-Instruct~\cite{bai2025qwen3}         & 235B  & 0.069 & 0.068 & 9.71 \\
& Gemini 3 Pro~\cite{google2025gemini3}                          & -     & 0.066 & 0.079 & 9.68 \\
\midrule
\multirow{11}{*}{SpecializedVLMs}
& Mistral OCR                         & -     & 0.164 & 0.144 & 8.84 \\
& DeepSeek-OCR~\cite{wei2025deepseek}                        & 3B    & 0.073 & 0.086 & 8.26 \\
& MonkeyOCR-3B~\cite{li2025monkeyocr}                        & 3B    & 0.075 & 0.129 & 9.27 \\
& OCRVerse~\cite{zhong2026ocrverse}                            & 4B    & 0.058 & 0.071 & --   \\
& MonkeyOCR-pro-3B~\cite{li2025monkeyocr}                    & 3B    & 0.075 & 0.128 & -    \\
& MinerU2.5~\cite{niu2025mineru2}                           & 1.2B  & 0.047 & 0.044 & -    \\
& PaddleOCR-VL~\cite{cui2025paddleocr}                        & 0.9B  & 0.035 & 0.043 & 9.51 \\
& HunyuanOCR~\cite{team2025hunyuanocr}                          & 0.9B  & 0.042 & -     & -    \\
& PaddleOCR-VL1.5~\cite{cui2026paddleocr}                     & 0.9B  & 0.035 & 0.042 & -    \\
& GLMOCR~\cite{duan2026glmocrtechnicalreport}                             & 0.9B  & 0.040 & 0.043 & -    \\
& dots.ocr~\cite{li2025dots}                            & 3B    & 0.048 & 0.053 & 9.29 \\
\rowcolor{gray!20}
& dots.mocr                        & 3B    & \textbf{0.031} & \textbf{0.029} & 9.54 \\
\bottomrule
\end{tabular}}
\label{tab:omnidocbench_pdfparse}
\end{table}

Metrics are collected from OmniDocBench v1.5~\cite{ouyang2025omnidocbench} and reported numbers in corresponding model publications. Results on pdf-parse-bench~\cite{horn2025benchmarking} are reproduced by Qwen3-VL-235B-A22B-Instruct under a unified evaluation setup. We omit OmniDocBench v1.5 Formula and Table metrics because they are highly sensitive to detection rules and ground-truth matching protocols, which may introduce non-negligible variance across implementations.

Overall, dots.mocr achieves the strongest performance on OmniDocBench v1.5 text transcription and reading-order metrics (TextEdit and ReadOrderEdit), indicating robust document parsing and ordering accuracy. In addition, we observe clear improvements over dots.ocr on formula-intensive pages, suggesting better recognition of mathematical content. In future work, we will prioritize boosting formula and table recognition performance via stronger high-quality supervision, targeted data scaling, and task-specific training that directly improves end-to-end recognition accuracy.

\subsection{Qualitative examples}
\label{sec:examples}

\begin{figure}[h]
    \centering
    \includegraphics[width=1.0\linewidth]{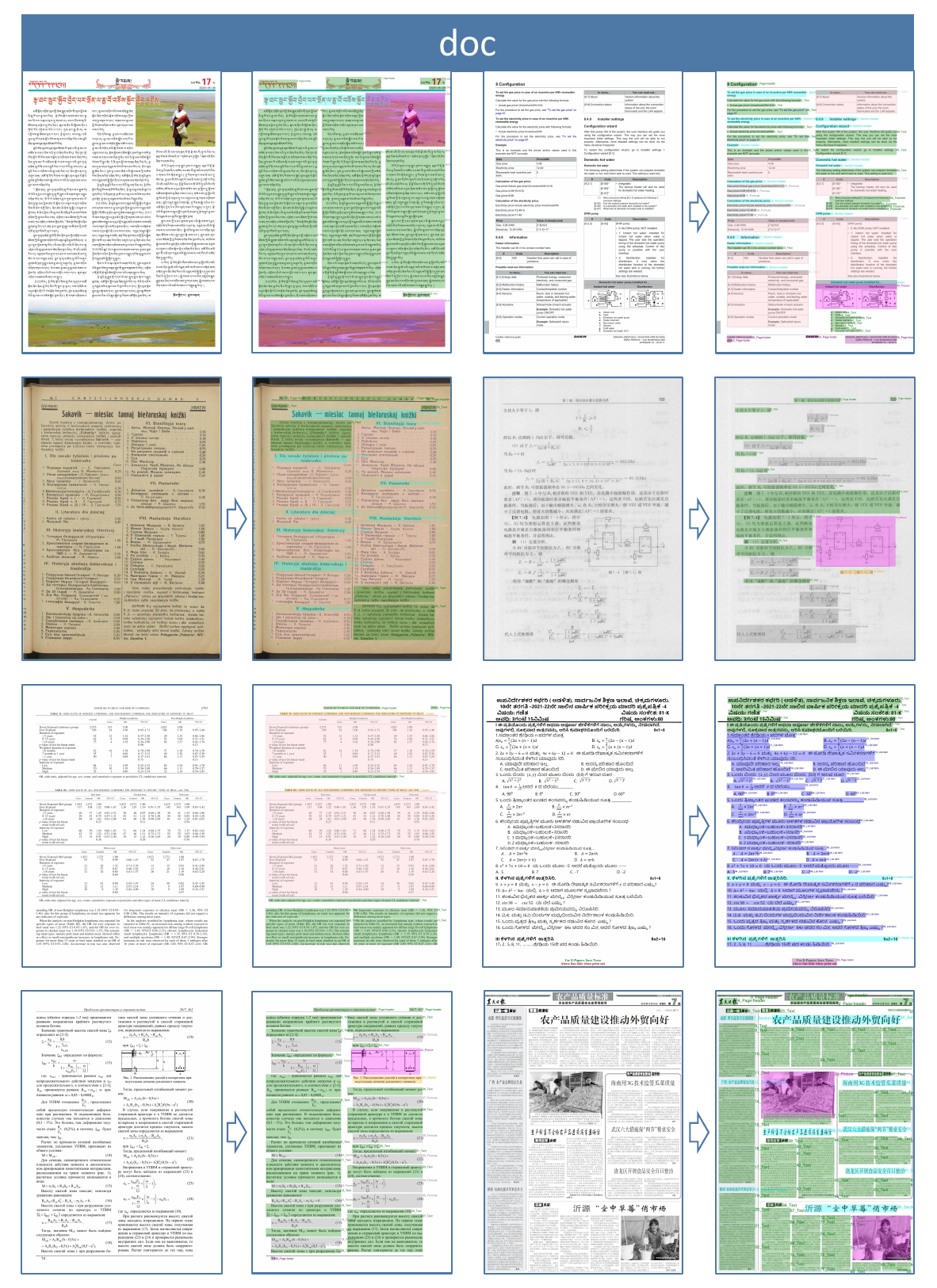}
    \caption{Qualitative layout analysis results of dots.mocr on heterogeneous documents. The model identifies and partitions structural elements such as titles, paragraphs, multi-column regions, dense tables, mathematical formulas, scanned text, and handwritten content, demonstrating robust document-level structural understanding across diverse formats and languages.}
    \label{fig:example-doc}
\end{figure}

\begin{figure}[h]
    \centering
    \includegraphics[width=0.72\linewidth]{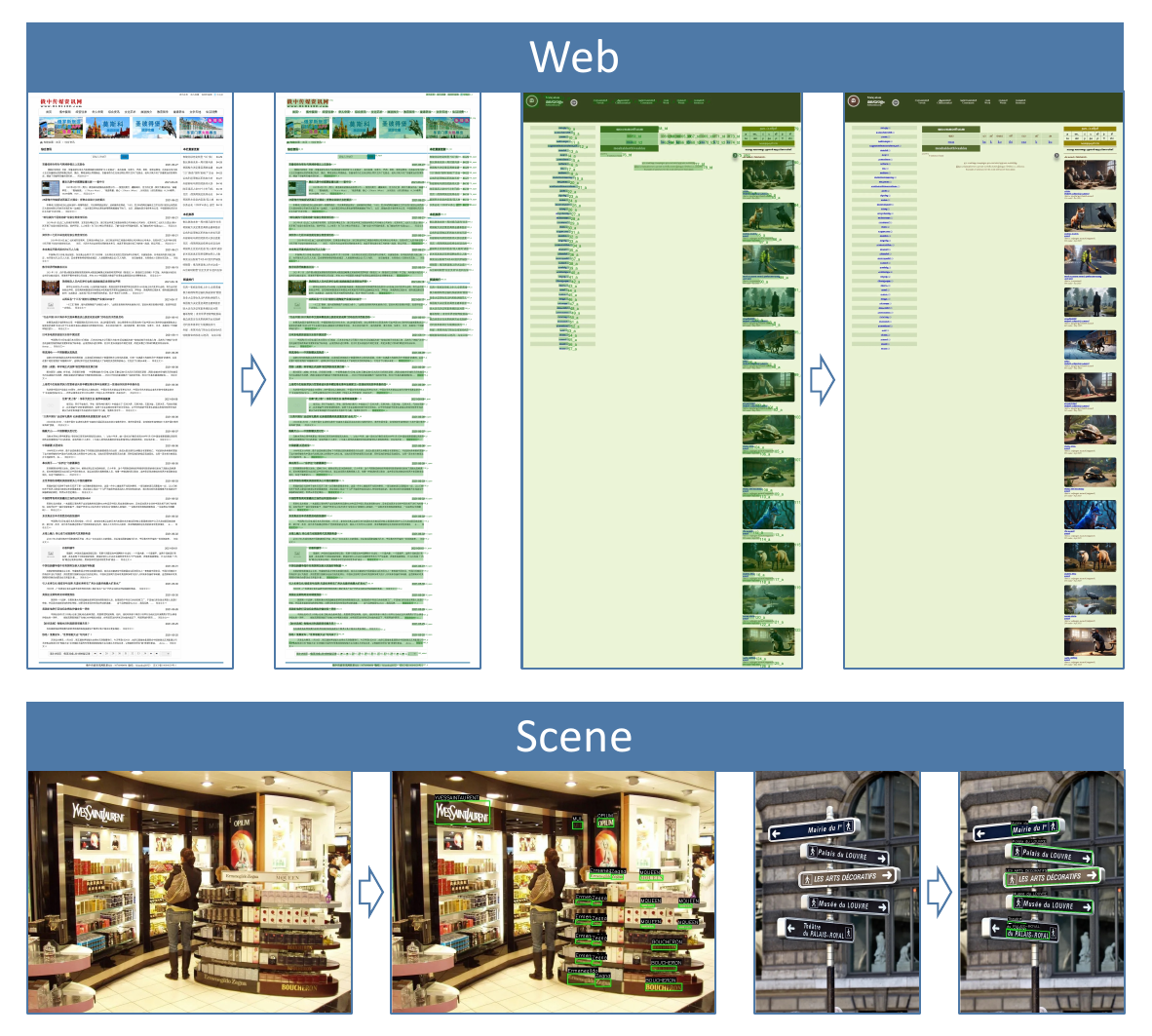}
    \caption{Parsing results of dots.mocr on full-length web screenshots and real-world scene images. The model preserves global reading order and structured components in long webpage layouts, while accurately recognizing and organizing scene text in complex real-world environments, highlighting generalization beyond conventional OCR benchmarks.}
    \label{fig:example-parse}
\end{figure}

\begin{figure}[h]
    \centering
    \includegraphics[width=0.72\linewidth]{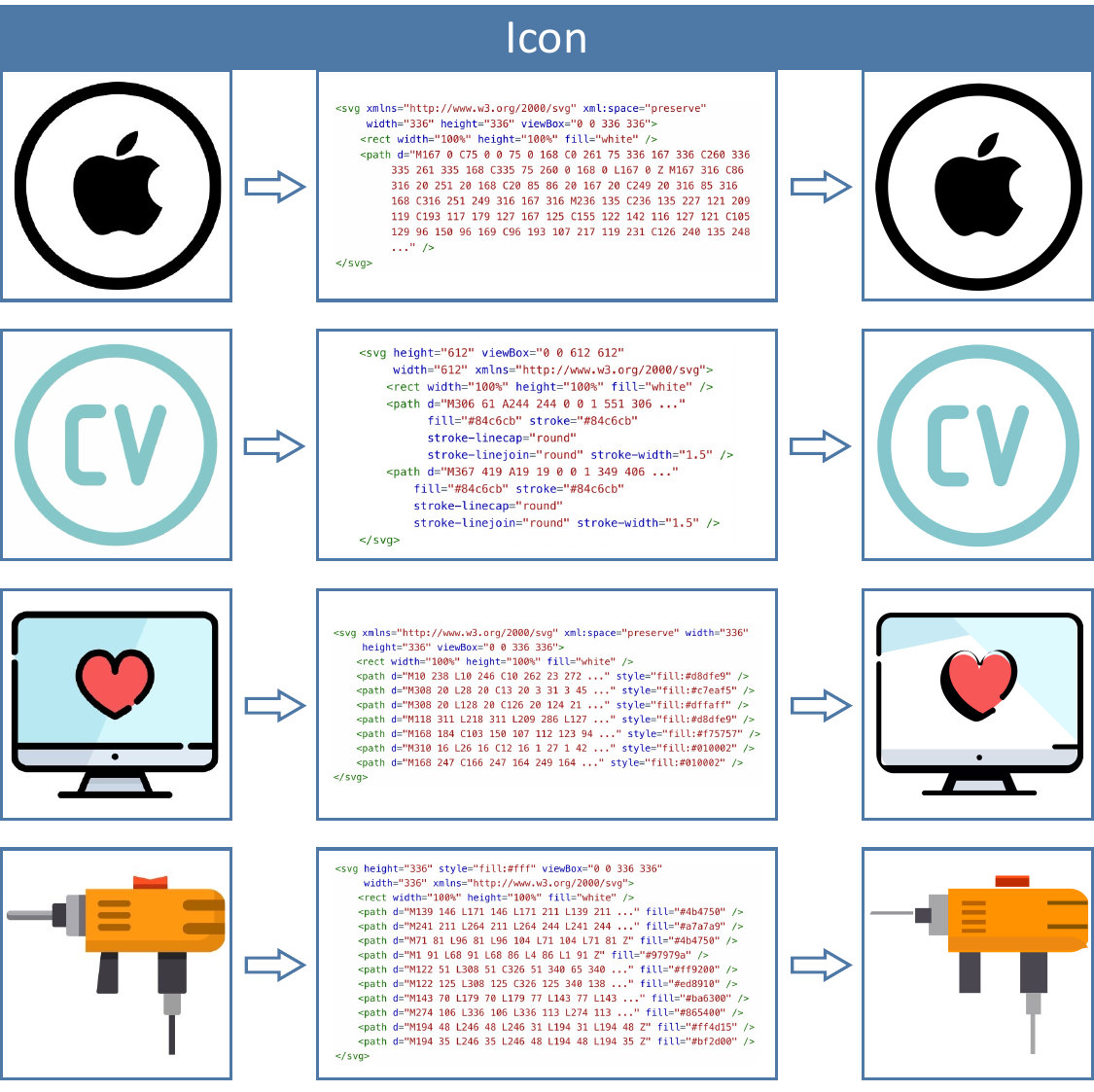}
    \caption{SVG parsing outputs and rendered reconstructions produced by dots.mocr-svg for icon images. The model converts raster icons into concise and executable SVG code, capturing geometric primitives, hierarchical grouping, and spatial relationships for scalable vector-level reconstruction.}
    \label{fig:example-icon}
\end{figure}

\begin{figure}[h]
    \centering
    \includegraphics[width=1.0\linewidth]{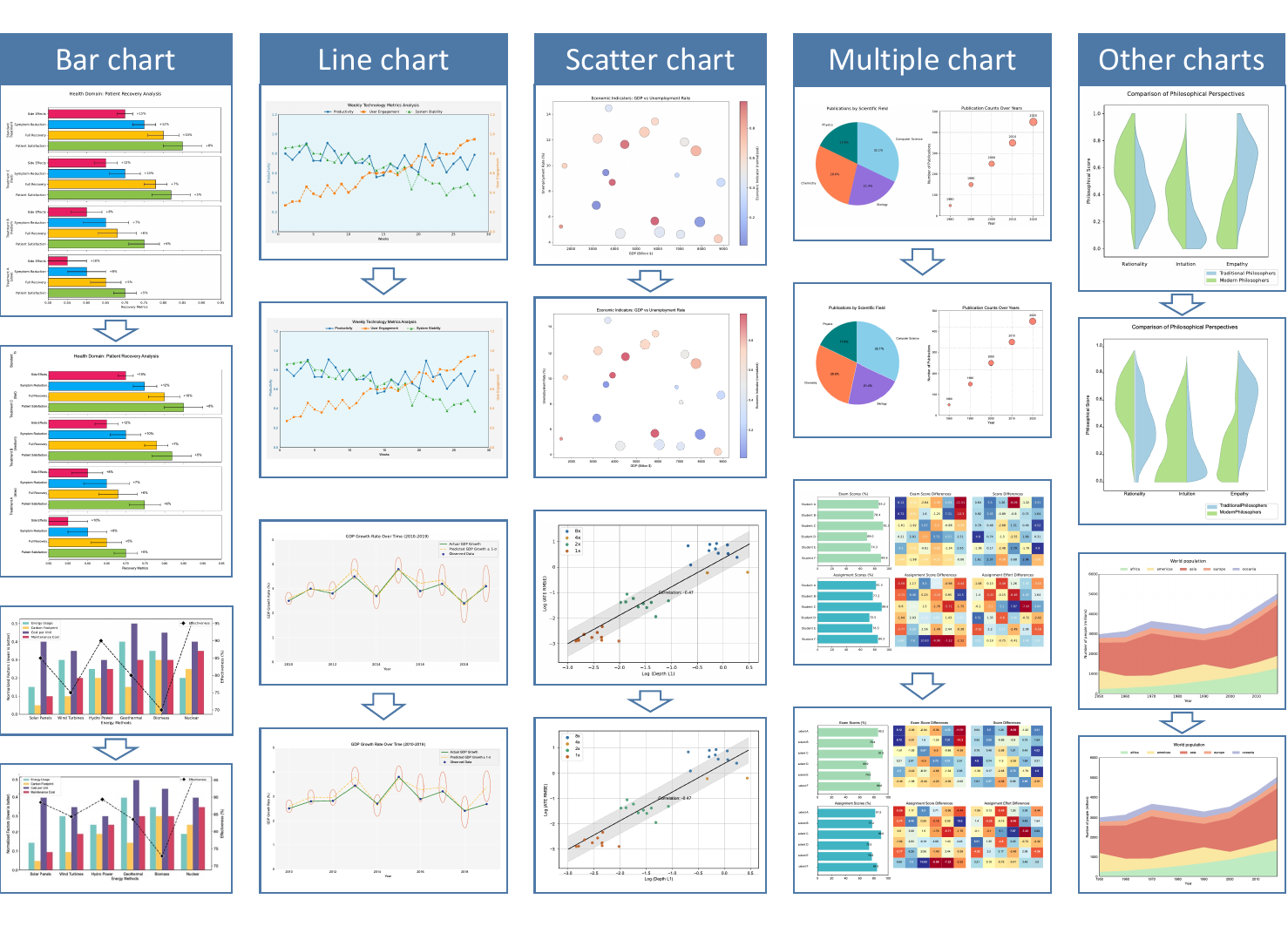}
    \caption{Rendered SVG reconstruction results of dots.mocr-svg for diverse statistical charts, including bar, line, scatter, and composite visualizations. The model recovers chart structure, axes, legends, data encodings, and semantic grouping through executable SVG representations, enabling editable and reusable vector outputs.}
    \label{fig:example-chart}
\end{figure}

\begin{figure}[h]
    \centering
    \includegraphics[width=1.0\linewidth]{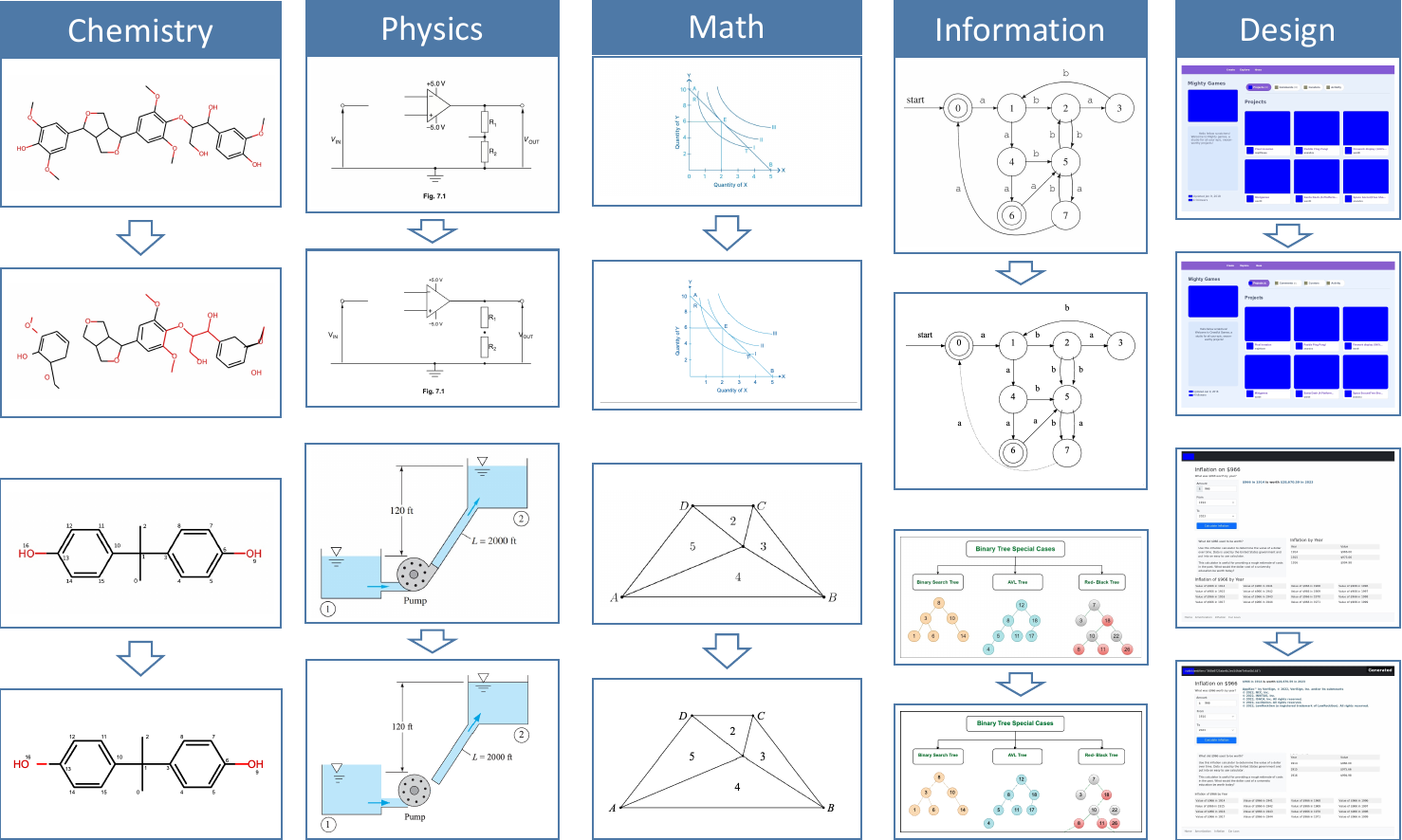}
    \caption{Rendered SVG reconstruction results of dots.mocr-svg for complex cross-disciplinary illustrations in science and design. The model captures intricate geometric structures, semantic components, and hierarchical organization, demonstrating its ability to generalize beyond standard charts to domain-specific structured graphics.}
    \label{fig:example-subject}
\end{figure}

\begin{figure}[h]
    \centering
    \includegraphics[width=1.0\linewidth]{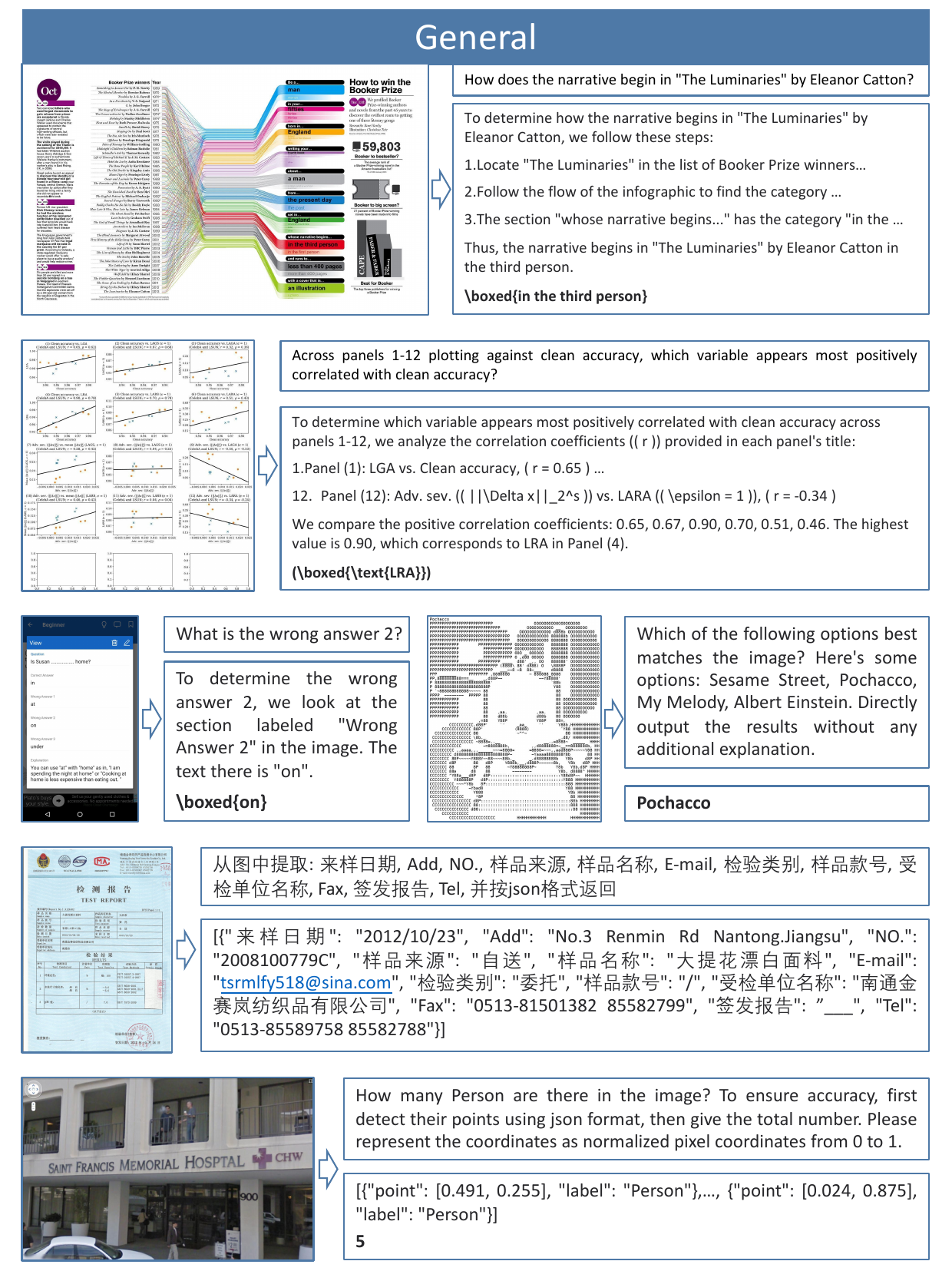}
    \caption{General-purpose visual question answering results of dots.mocr. While retaining strong document parsing and SVG reconstruction capabilities, the model produces coherent and context-aware responses across documents, charts, UI screenshots, and complex illustrations, indicating competitive performance in broader vision–language tasks.}
    \label{fig:example-general}
\end{figure}

This section presents qualitative examples of the outputs produced by dots.mocr and dots.mocr-svg, shown in  Figures~\ref{fig:example-doc}--\ref{fig:example-general}.

\subsection{OCR Arena}
\begin{figure}
    \centering
    \includegraphics[width=1.0\linewidth]{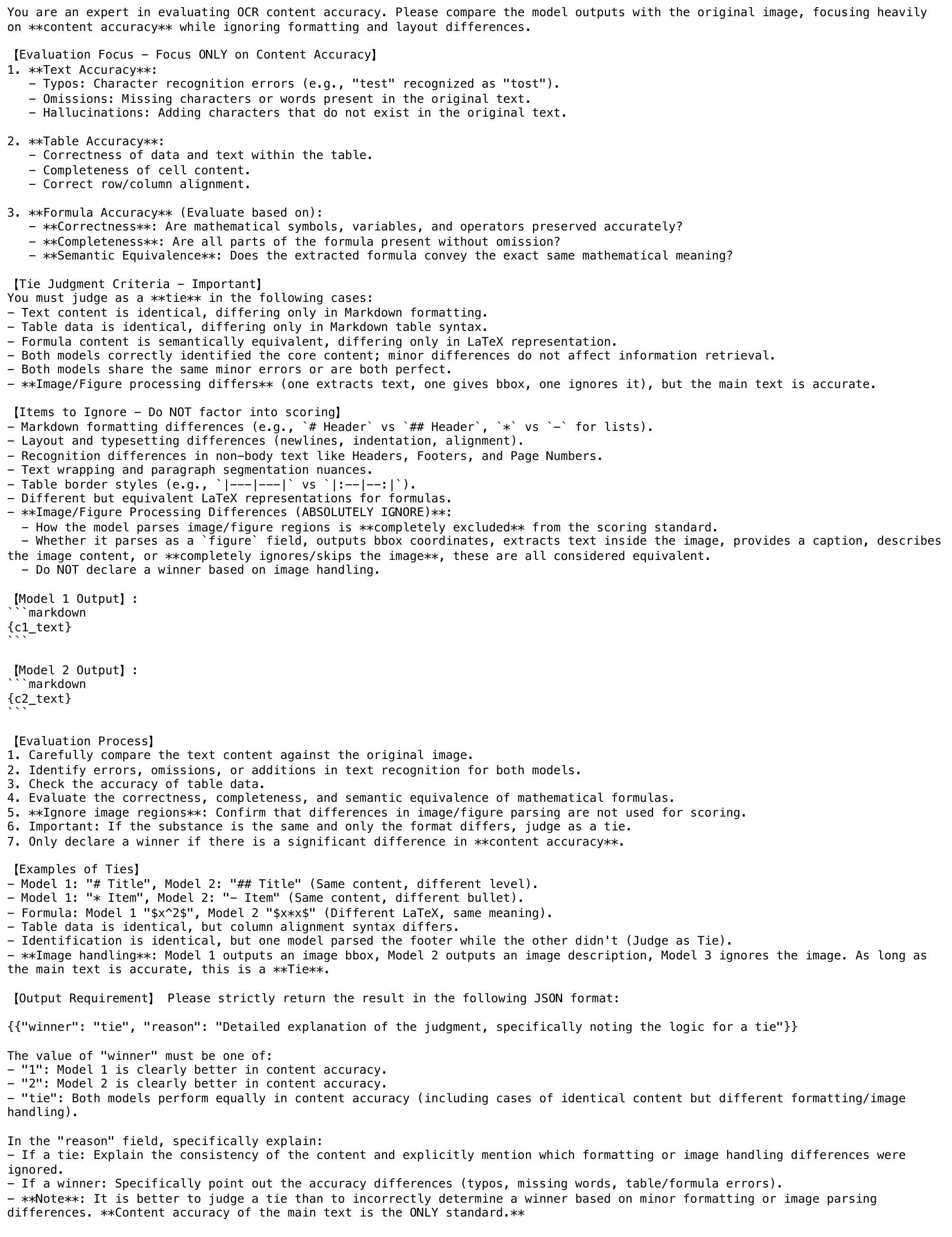}
    \caption{Prompt used in OCR Arena}
    \label{fig:prompt}
\end{figure}

\begin{figure}
    \centering
    \includegraphics[width=1.0\linewidth]{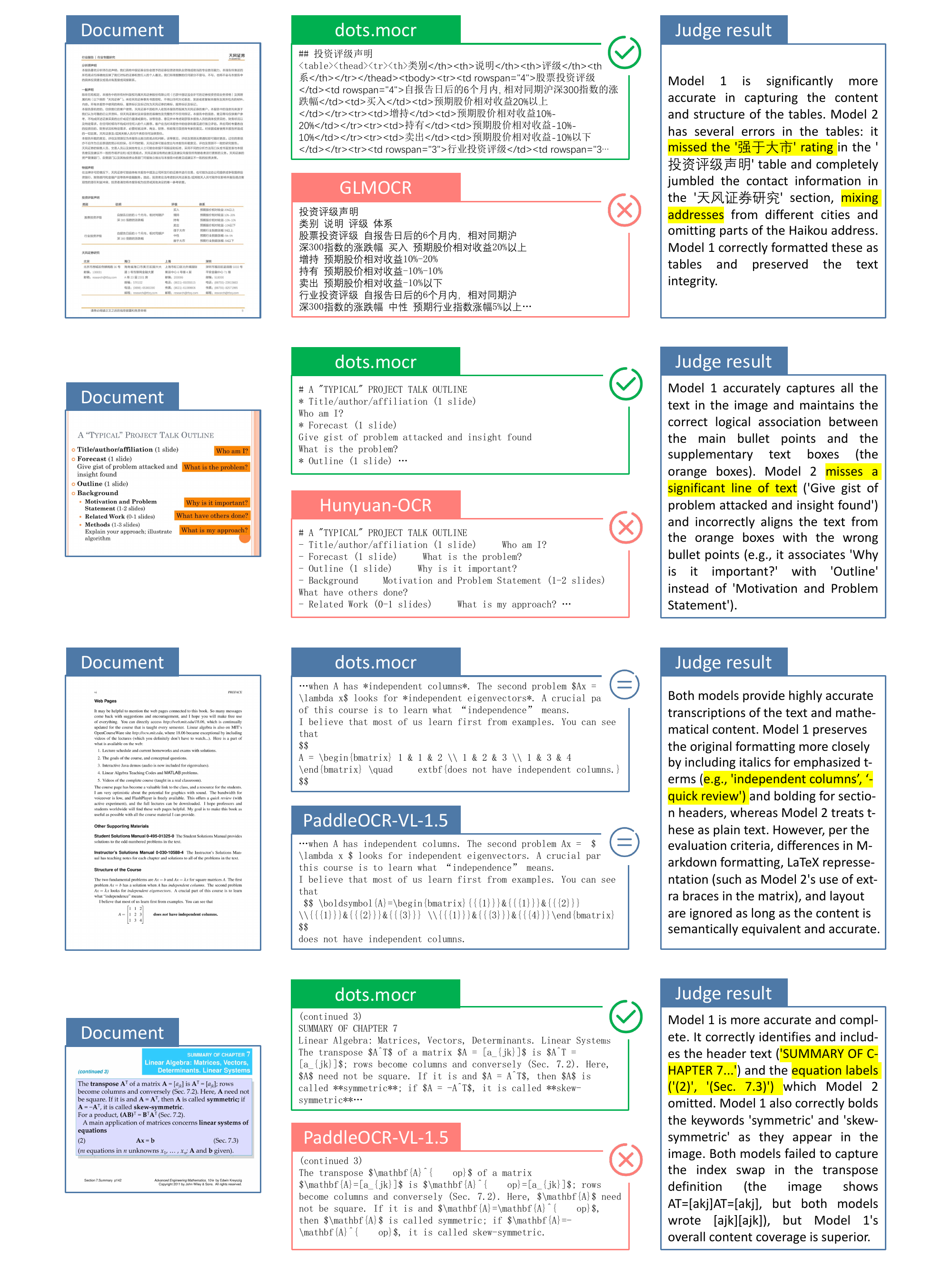}
    \caption{Example judging results in OCR Arena, showing document images, OCR outputs from two models, and the explanations produced by the VLM judge.}
    \label{fig:judge_results}
\end{figure}

We adopt a high-capacity vision-language model (VLM) as an impartial judge (e.g., Gemini 3 Flash) to compare two Markdown OCR outputs conditioned on the same source document image. The judging prompt is shown in Figure~\ref{fig:prompt} and some judging results are provided in Figure~\ref{fig:judge_results}.

\end{document}